\documentclass[letterpaper]{article} 
\usepackage{aaai24}  
\usepackage{times}  
\usepackage{helvet}  
\usepackage{courier}  
\usepackage[hyphens]{url}  
\usepackage{graphicx} 
\usepackage{natbib}  
\usepackage{caption} 
\usepackage{amsmath}
\usepackage{booktabs}       
\usepackage{amsfonts}       
\usepackage{nicefrac}       
\usepackage{microtype}      
\usepackage{xcolor}         
\usepackage{newfloat}
\usepackage{listings}
\usepackage{multirow}
\usepackage{multicol}
\usepackage{mathtools}
\usepackage{amsmath}
\usepackage{bm}
\usepackage{graphicx}
\usepackage{subcaption}
\usepackage{enumitem}
\usepackage{amsthm}
\usepackage{amssymb}
\usepackage{booktabs}
\usepackage{tcolorbox}
\usepackage{algorithm}
\usepackage{algorithmic}
\usepackage{amssymb}
\usepackage{graphicx}
\usepackage{subcaption}
\usepackage{booktabs}
\usepackage{tabularx}
\usepackage{subcaption}
\newcommand{\model}{IGM}
\usepackage{newfloat}
\usepackage{listings}
\DeclareCaptionStyle{ruled}{labelfont=normalfont,labelsep=colon,strut=off} 
\floatstyle{ruled}
\newfloat{listing}{tb}{lst}{}
\floatname{listing}{Listing}
\title{Graph Invariant Learning with Subgraph Co-mixup for \\Out-Of-Distribution Generalization}
\author{
    Tianrui Jia\textsuperscript{\rm 1},
    Haoyang Li\textsuperscript{\rm 2},
    Cheng Yang\textsuperscript{\rm 1}$^*$,
    Tao Tao\textsuperscript{\rm 3},
     Chuan Shi\textsuperscript{\rm 1}\thanks{Corresponding authors.}
}
\affiliations{
    \textsuperscript{\rm 1}Beijing University of Posts and Telecommunications\\
    \textsuperscript{\rm 2}Tsinghua University\\
    \textsuperscript{\rm 3}China Mobile Information Technology Co. Ltd. \\

    \{jiatianrui, yangcheng, shichuan\}@bupt.edu.cn, lihy18@mails.tsinghua.edu.cn, taotao@chinamobile.com
}

\usepackage{bibentry}

\begin{document}

\maketitle

\begin{abstract}
Graph neural networks (GNNs) have been demonstrated to perform well in graph representation learning, but always lacking in generalization capability when tackling out-of-distribution (OOD) data. Graph invariant learning methods, backed by the invariance principle among defined multiple environments, have shown effectiveness in dealing with this issue. However, existing methods heavily rely on well-predefined or accurately generated environment partitions, which are hard to be obtained in practice, leading to sub-optimal OOD generalization performances.
In this paper, we propose a novel graph invariant learning method based on invariant and variant patterns co-mixup strategy, which is capable of jointly generating mixed multiple environments and capturing invariant patterns from the mixed graph data. Specifically, we first adopt a subgraph extractor to identify invariant subgraphs. Subsequently, we design one novel co-mixup strategy, i.e., jointly conducting environment Mixup and invariant Mixup. For the environment Mixup, we mix the variant environment-related subgraphs so as to generate sufficiently diverse multiple environments, which is important to guarantee the quality of the graph invariant learning. For the invariant Mixup, we mix the invariant subgraphs, further encouraging to capture invariant patterns behind graphs while getting rid of spurious correlations for OOD generalization. We demonstrate that the proposed environment Mixup and invariant Mixup can mutually promote each other.
Extensive experiments on both synthetic and real-world datasets demonstrate that our method significantly outperforms state-of-the-art under various distribution shifts\footnote{Code available at https://github.com/BUPT-GAMMA/IGM}.
\end{abstract}

\section{Introduction}

Graph data is ubiquitous in the real world, such as molecular networks, protein networks, social networks. Graph representation learning ~\cite{chen2020graph,hamilton2017representation} achieves deep learning on graphs by encoding them into vectors in a latent space. 
Graph neural networks (GNNs)~\cite{gcn,gin,gat,graphsage}, as one of the most popular graph representation learning methods, have attracted wide attention in the last decade.~\cite{lee2018graph, gin}. 

Despite their noticeable success, existing GNNs heavily rely on the identically distributed (I.D.) assumption~\cite{erm}, i.e., the training and test data are sampled from an identical distribution. However, various forms of distribution shifts between the training and testing datasets widely exist in the real world, since the uncontrollable data generation mechanisms, resulting in OOD~\cite{hu2020open,ji2022drugood,koh2021wilds} scenarios. 
For instance, in graph classification tasks, there could be significant distribution shifts existing in graph size~\cite{bevilacqua2021size,yehudai2021local}, node degree~\cite{yoo2023disentangling}, and structure (e.g., molecule scaffold)~\cite{ji2022drugood} between the training and testing graphs. 
Existing GNNs that perform well on the training data by capturing the spurious correlations significantly fail to generalize to OOD testing graph data. 
Therefore, it is of paramount importance to capture the invariant relationships between predictive graph patterns and labels.

Invariant learning~\cite{arjovsky2019invariant,krueger2021out,creager2021environment,ahuja2021invariance} emerges as a prevalent strategy for tackling the challenge of generalization to OOD data. The basic assumption of invariant learning method is the invariance principle among defined multiple environments, namely there existing a proportion of input data capturing invariant relations with the labels across distinct environments ~\cite{arjovsky2019invariant}. Consequently, a predictor that performs well across multiple pre-defined environments is guaranteed to possess generalization capabilities for unseen data distributions~\cite{arjovsky2019invariant}. 
In the field of graphs, existing graph invariant learning methods~\cite{wu2022discovering,miao2022interpretable,li2022learning,chen2022learning,yang2022learning} consider that within each environment the graph data can be decomposed into two components, including invariant subgraphs that have deterministic and truly predictive relations with the labels, and environment subgraphs that could exhibit spurious correlations with the labels. 
The main goal of them is focused on obtaining diverse training environments. For example,
DIR~\cite{wu2022discovering} generates multiple training environments for invariant learning by implementing distribution interventions on graphs, while GIL~\cite{li2022learning} clusters environment subgraphs and treats each cluster as an environment. The performance of invariant learning heavily relies on the diversity of environment partitioning. In other words, if different environments are not diverse, these methods will not sufficiently get rid of spurious correlations, showing poor OOD generalization ability.
For example, when there exists size distribution shift between the training set and the test set, suppose the size of the graphs in the training set is 6-8, while the size of the graphs in the test set is 30-50, then no matter how environments are partitioned within the training set, it would be challenging for the model to demonstrate satisfactory generalization capabilities in the test set.
However, the environments generated by existing graph invariant learning methods cannot possess sufficient distribution shifts among different environments. For DIR~\cite{wu2022discovering}, although it is theoretically possible to mitigate the influence of the environment, all environments are relatively similar in the initial stages of training, performing suboptimally in practice since environments can not be proved to be diverse~\cite{chen2022pareto}. 
Similarly, for GIL~\cite{li2022learning}, if the training set itself does not have significantly diverse latent environments, the generated environments during the training process will also not be enough to learn invariant patterns.
In addition to these two representative methods, existing methods are generally hard to achieve a diverse environment partitioning, resulting in suboptimal performance and largely hindering the OOD generalization.

To tackle these problems, in this paper, we are the first to study mixup-based graph invariant learning for graph OOD generalization, to the best of our knowledge.
Although Mixup~\cite{zhang2017mixup} and their variations~\cite{verma2019manifold,chou2020remix,kim2021co,yun2019cutmix}, as one type of interpolation-based data augmentation methods that amalgamate two training instances and the labels to generate new instances are proposed in the literature,
the existing graph mixup methods~\cite{han2022g,wang2021mixup,park2022graph,guo2021ifmixup} are only based on mixing up entire graphs, which can definitely introduce spurious correlations since they do not explicitly distinguish invariant and environment subgraphs during conducting mixup,  so as to degrade the model's generalization performance on OOD graph data.
Incorporating mixup with invariant learning for graph out-of-distribution generalization is promising but poses great challenges as follows and has not been explored:
\begin{itemize}
    \item How to design mixup to generate diverse enough environments which have enough distribution shifts for invariant learning.
    \item How to improve the mixup method so that the mixed-up graph data only retains invariant information while excluding environmental-related spurious correlations for OOD generalization.
\end{itemize}

To address the aforementioned challenges, we propose a novel graph invariant learning method based on invariant and variant co-mixup strategy, herein referred to as \textbf{I}nvairnt learning on \textbf{G}raph with co-\textbf{M}ixup (\model). Firstly, we design an invariant subgraph extractor to identify the invariant subgraphs and consider their complements as the environment-related environment subgraphs. Then, we design an environment Mixup module based on the environment subgraphs to encourage the generated environments that are sufficiently diverse for graph invariant learning. 
We generate a variety of environments by concatenating invariant and environment subgraphs with different labels. The environments generated in this tailored way will have sufficient distribution shifts so as to be diverse enough. Next, in order to ensure that the mixed graph data only retains invariant information, we design an invariant Mixup module to perform mixup only on invariant subgraphs rather than the whole graphs. 
Performing invariant and environment subgraph co-mixup with these two modules above can effectively get rid of spurious correlations from the entire graph.
More importantly, we also show that environment Mixup and invariant Mixup modules of the co-mixup strategy can mutually promote each other, for the promising performances of the OOD generalization capabilities.

We conducted extensive experiments on three artificially synthesized datasets and nine real-world datasets to verify the effectiveness of our proposed method for various types of distribution shifts. Compared to the state-of-the-art baselines, our method shows significant improvements, e.g., an average of 7.4\% improvement on real-world datasets. Furthermore, we have verified the effectiveness of each module and performed visualization experiments on the learned invariant subgraphs to conduct deeper analyses.
Our contributions can be summarized as follows: (1) We design an invariant and environment subgraph co-mixup based graph invariant learning method for OOD generalization. To the best of our knowledge, this is the first work to automatically generate enough diverse environments for graph invariant learning.
(2) We design an environment Mixup module to generate environments which have enough distribution shifts, leading to better invariant learning on graphs. 
(3) We propose an invariant Mixup method to encourage the mixed-up data only retain invariant graph patterns. This novel design mitigates the impact of spurious correlations in the whole graph. We demonstrate that our designed environment Mixup and invariant Mixup can mutually promote each other in practice, thereby enhancing the generalization capability on OOD graph data. 
(4) We conduct extensive experiments on both synthetic and real-world datasets to show that our proposed method has the most competitive OOD generalization ability via significantly outperforming state-of-the-art on various types of distribution shifts.

\section{Related Works}

\paragraph{Graph Neural Network.}
Graph Neural Networks (GNNs)~\cite{gcn,graphsage,gin,gat} aggregate the neighbors of nodes through a message-passing mechanism to obtain individual node representations. Subsequently, a pooling function is employed to derive a global graph representation, which is then utilized for subsequent classification tasks.
Inspired by the spectral method~\cite{BrunaZSL13,DefferrardBV16}, GNN is designed to use convolutional neural networks to aggregate neighbors’ features~\cite{gcn,graphsage}. Due to the good performance of the attention mechanism, attention is introduced to GNN, which is known for GAT~\cite{gat}.
However, traditional GNNs fail to achieve generalization on OOD data.
\paragraph{OOD Generalization on Graphs.}
Currently, graph OOD generalization methods~\cite{xia2023learning,miao2022interpretable,li2022learning,wu2022discovering,yang2022learning,liu2022graph,sui2022causal,buffelli2022sizeshiftreg,zhang2022dynamic,chen2022ba,li2022ood} can be primarily categorized into two approaches. The first, based on information bottleneck methods such as CIGA~\cite{chen2022learning} and GSAT~\cite{miao2022interpretable}, achieves generalization by maximizing mutual information between labels and invariant subgraphs while minimizing mutual information between the subgraph and the entire graph. The second approach, based on invariant learning methods~\cite{arjovsky2019invariant,krueger2021out,creager2021environment,ahuja2021invariance}, like DIR~\cite{wu2022discovering} and GIL~\cite{li2022learning}, defines environments within datasets and incorporates a regularization term between these environments. This method aims to learn cross-environment invariant information, thereby facilitating OOD generalization.
\section{Notations and Preliminaries}

\paragraph{Notations.} 
Denote a graph dataset as $\mathcal{G} = \{G_i,Y_i\}_{i=1}^N$.
Due to the uncontrollable data generation mechanism~\cite{bengio2019meta}, we follow the literature~\cite{arjovsky2019invariant,ahuja2021invariance} to consider realistic yet challenging scenarios that there exist unobservable distribution shifts between training and test sets P(G$_{train}) \neq$ P(G$_{test})$ since the training and test graph data are sampled from different environments.
The label space of graphs and labels are $\Bbb{G}$, $\Bbb{Y}$.

\paragraph {Problem Formulation.}

Following existing works~\cite{wang2021mixup,li2022learning}, we assume each graph $G_i$ consists of two parts, namely invariant subgraph $G_i^I$ and environment subgraph $G_i^E$, where $G_i^E$ is the complement of $G_i^I$. Denote the invariant subgraph set as $\mathcal{G}^I = \{G_i^I,Y_i\}_{i=1}^N$ and environment subgraph set as $\mathcal{G}^E = \{G_i^E,Y_i\}_{i=1}^N$.
We use the subscript to denote the corresponding train and test set, i.e., $\mathcal{G}^I_{train}$,$\mathcal{G}^E_{train}$,$\mathcal{G}^I_{test}$,$\mathcal{G}^E_{test}$.
$G_i^I$ determine its label $Y_i$ so that it has invariant relations with the label and should be captured for OOD generalization, i.e., P(Y$_{train}|$G$_{train}^I$)$ =$ P(Y$_{test}|$G$_{test}^I$) $=$ P(Y$|$G$^I$), where G$_{train}^I$, G$_{test}^I$, G$^I$ denotes the random variables for $\mathcal{G}^I_{train}$, $\mathcal{G}^I_{test}$, $\mathcal{G}^I$.
In contrast, $G_i^E$ contains information which has spurious relation with $Y_i$ so that it has variant relations with the label and should be got rid of for stable performances among different environments.

Thus, our objective is to identify the invariant subgraphs within the graph and only use them to make OOD generalized predictions. By extracting the right invariant subgraph of each graph, our model will generalize well in testing environment~\cite{li2022learning,chen2020graph,wu2022discovering}.

\section{Methodology}
\begin{figure*}
\centering
\includegraphics[width=0.8\textwidth]{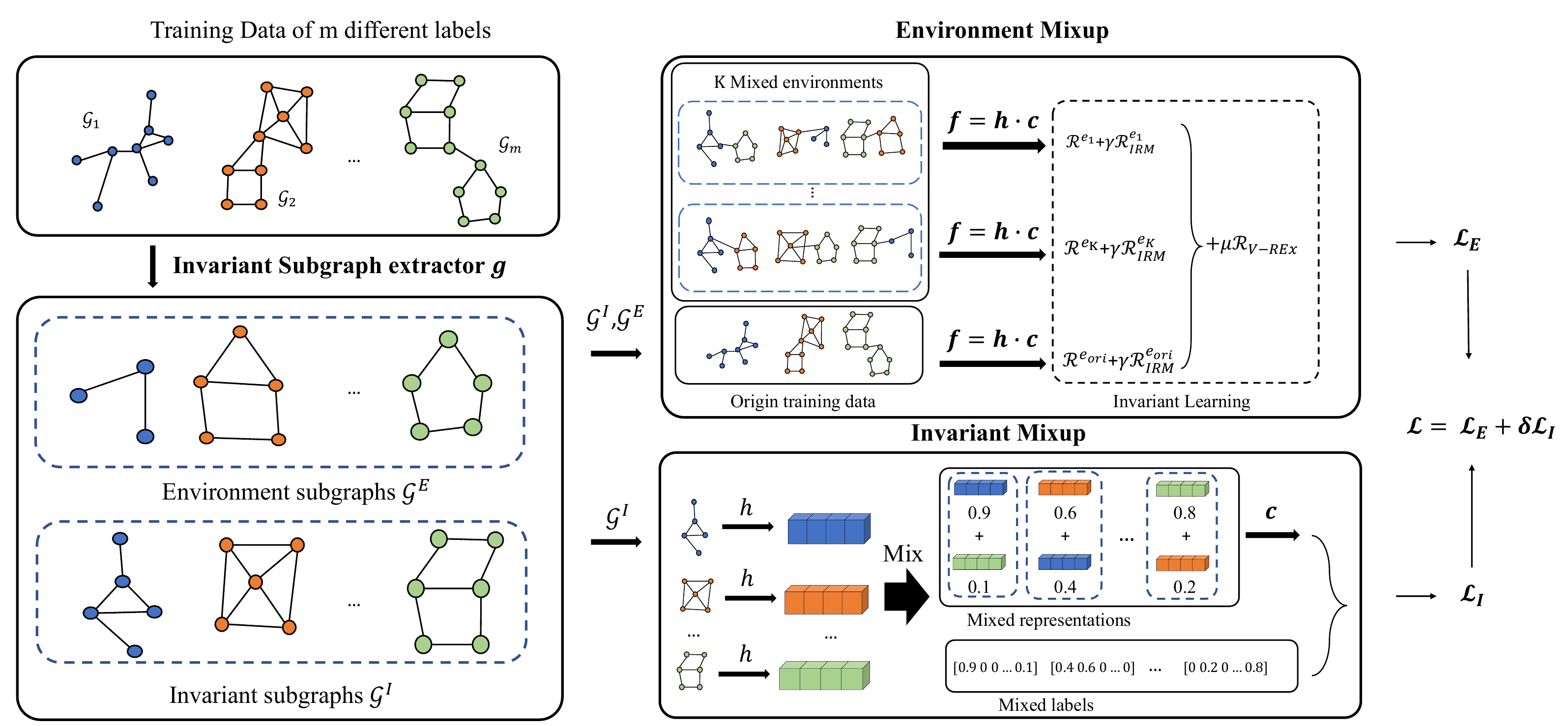}
\caption{The overall framework of our proposed IGM. Invariant subgraph extractor $g$ splits each graph into invariant subgraph and environment subgraph. Following this, we employ two mixup strategies: (1) Concatenating invariant subgraphs and environment subgraphs from different labels to generate $K$ new environments, upon which we conduct invariant learning. (2) Mixing invariant subgraphs from different labels to augment the data. Here, $h$ represents a GNN used for feature extraction, and $c$ denotes an MLP utilized for classification.}
\label{fig:framework}
\end{figure*}

In this section, we first present the overall framework of our IGM. Then we will introduce our invariant subgraph extractor. Finally, we will describe the two mixup modules based on the extracted subgraphs, namely environment Mixup and invariant Mixup,   
An overview of IGM is shown in Figure~\ref{fig:framework}.

\subsection{Overall Framework}
To tackle the limitations of the existing graph invariant learning methods' strong dependence on the predefined environment partitions, we propose to incorporate mixup~\cite{zhang2017mixup} and invariant learning to generate mixed environments and capture invariant patterns from mixed graphs simultaneously. Given the input data, we first use an invariant subgraph extractor to extract the invariant and environment subgraphs from each graph. Subsequently, we apply environment Mixup and invariant Mixup to update the parameters of the invariant subgraph extractor.

Specifically, the environment Mixup module is designed to generate environments with sufficient distribution shifts and the invariant Mixup module is proposed to prevent spurious correlations within the graph from affecting the mixup. Note that these two modules can mutually enhance each other's learning: on the one hand, the environment Mixup module is able to partition environments with sufficient distribution shifts, thereby facilitating the invariant Mixup to capture more invariant information. On the other hand, as the invariant Mixup captures more invariant information, it can further aid the environment Mixup in achieving a more refined environmental partition, subsequently promoting the invariant learning of the environment Mixup.

\subsection{Invariant Subgraph Extractor}

We use $g$ to represent the subgraph extractor, $G_{i}^I = g(G_i)$, corresponding to the invariant feature extractor $g$ in the previous section. The idealized invariant subgraph extractor $g^*(\cdot)$ should satisfy:
\begin{equation}
\begin{split}
\text{{P}}_{e_{1}}(\text{{Y}}|g^*(\text{{G}})) = \text{{P}}_{e_{2}}(\text{{Y}}|g^*(\text{{G}}))\quad,  \forall e_1,e_2 \in \mathcal{E}, \\
\mathcal{R}(f\circ g^*(\text{{G}})) = \min\mathcal{R}(f\circ g(\text{{G}})),\qquad
\end{split}
\label{extractor1}
\end{equation}

\noindent where $\mathcal{E}$ is the set of environments. $\mathcal{R}(\cdot)$ is the risk function that can be a cross-entropy, and $f$ represents the classifier.  

Now we instantiate $g$ with learnable parameters. For a given graph $G$, its nodes set and edges set are $V_G$ and $E_G$ respectively. 
The $p_{(u,v)}$ represents the probability that edge $(u,v)$ is selected as an edge in the invariant subgraph $G_I$, and we get it via a GNN$_{enc}$ and an MLP$_{enc}$:


\begin{equation}
\begin{split}
\Omega = \text{GNN}_{enc}(G),\qquad\\
\phi_{(u,v)} = \Omega_u \parallel \Omega_v,\qquad\\
p_{(u,v)} = \text{MLP}_{enc}(\phi_{(u,v)}),
\end{split}
\label{extractor2}
\end{equation}
\noindent where $\Omega$ is the nodes representations, $\Omega_u$, $\Omega_v$ is the representation of node $u$, $v$.

 Next, we sample edges based on distribution $\xi_{uv} \sim Bern(p_{uv})$ to get $G_I$. Due to the non-differentiability of this sampling process, we employ the Gumbel Softmax~\cite{jang2016categorical} technique to make it differentiable. 

In practice, we set a maximum ratio $r$ to avoid the extracted subgraph being overly large, as defined by the condition $\frac{|V_{G_I}|}{|V_{G}|}\leq r$. With our subgraph extractor, we can adaptively select edges instead of selecting a fixed ratio of nodes or edges. In the subsequent experimental section, we will report the values of $r$.

\subsection{Mixup for Out-of-Distribution Generalization}

Following the extraction of the invariant and environment subgraphs, we proceed to apply two distinct types of mixup to them, namely Invariant Mixup and Environment Mixup.

\paragraph{Environment Mixup.}

Here we design an Environment Mixup method to generate diverse enough environments which have sufficient distribution shift with origin data, which makes invariant learning on graphs more effective. Furthermore, the Environment Mixup enhances the Invariant Mixup, which will be mentioned in the following section. 

Let $L:\Bbb{Y} \rightarrow \Bbb{Y}$ be a mapping function that maps a label to a different one. For each $G_{i}^I \in \mathcal{G}^I_{train}$, we random select an $G_{j}^E \in \mathcal{G}^E_{train}$, whose label $Y_j = L(Y_i)$. Then we mixup the two subgraphs by randomly adding edges between the two graphs according to the node degree. The number of edges added is $n^{G_{i}^I G_{j}^E}_{add} = r_{add}(| E_{G_{i}^I} |+| E_{G_{j}^E} |)$, and $r_{add}$ is a pre-defined ratio. Since the invariant subgraphs determine the label, we define the label of the augmented graph as $G_{i}^I$'s label $Y_i$. We can obtain multiple label mapping functions and augmentations for data with multiple classes. For $K$ augmentations, denotes the $k$-th label function as $L_k$. We define the $k$-th augmentation as $\mathcal{G}_{aug}^{k}$:

\begin{equation}
\begin{split}
\mathcal{G}_{aug}^{k} = \{\text{{Mix}}(G_{i}^I,G_{j}^E)|G_{i}^I \in \mathcal{G}^I_{train}, G_{j}^E \in \mathcal{G}^E_{train},\\
Y_j = L_k(Y_i),i = 1,2,3...,|\mathcal{G}^I_{train}| 
\}.
\end{split}
\end{equation}

Since $G_{j}^E$ is only spuriously related with $Y_j$, graph $G_{i}^I$  concatenated with $G_{j}^E$ is more different than that with its own spurious subgraph $G_{i}^E$ or with a same substructure. So we consider each augmentation $\mathcal{G}_{aug}^{k}$ as an environment that has obvious distribution shifts with the origin training set.

After obtaining $K$ augmented environments, we adopt invariant learning on them to enable our model to learn invariant information across environments and extract correct invariant subgraphs that satisfy previous assumptions.

Drawing from the invariant learning literature~\cite{chen2022pareto}, combining different invariant regularizers can improve the generalization ability of modes further. During our training of Environment Mixup, V-REx~\cite{krueger2021out} regularizer is less impactful initially due to similar environments, while the IRM~\cite{arjovsky2019invariant} contributes more to the optimization procedure. But as spurious correlations increase in later stages, the effectiveness of IRM regularizer reduces, while V-REx gains importance. 
Hence, using both regularizers together leads to a better ability of generalization. We then formulate the overall risk following the IRMX~\cite{chen2022pareto} literature:
\begin{align}
\mathcal{L}_{E} = \sum_{e=0}^{K} &\Big(\mathcal{R}^e(f) + \gamma \mathcal{R}_{\text{IRM}}^e\Big) + \mu \mathcal{R}_{\text{V-REx}},
\label{nc_mix}
\end{align}
\noindent where $K$ is the number of environments, and $e=0$ represents the orignal data. $f$ is the classifier, and $\mathcal{R}^e(f)$ represents CrossEntropy loss on environment $e$.
$\gamma$ is the weight of  IRM regularizer $\mathcal{R}_{\text{IRM}}^e = \parallel \bigtriangledown_{w|w=1.0}\mathcal{R}^e(w \cdot f) \parallel^2$, and $\mu$ is the weight of of V-REx regualrizer $\mathcal{R}_{\text{V-REx}} = \text{Var}_e(\mathcal{R}^e(f))$. Var($\cdot$) denotes the variance of risks over the environments.
We instantiate $f$ with GNN$_{fea}$ and MLP$_{cls}$ as follows:
\begin{equation}
\begin{split}
\Psi = \text{{GNN}}_{fea}(G), \psi = \text{{Pooling}}(\Psi),\\
\hat{Y} = \text{{SoftMax}}(\text{{MLP}}_{cls}(\psi)),\quad
\end{split}
\label{e-mix1}
\end{equation}
\noindent where $\Psi$ is the node representation of $G$ and Pooling is the readout function. For clarity of presentation, we denote Pooling(GNN$_{fea}(G)$) as $h$, MLP$_{cls}$ as $c$ in Figure \ref{fig:framework}.

In the previous discussions, mixup was primarily utilized as a data augmentation technique to promote invariant learning in different environments. However, mixup also serves as a good regularizer for improving model generalization. In the following section, we will present how to leverage the extracted invariant subgraphs to perform mixup, thereby further enhancing the generalization capability of the model on out-of-distribution data.

\paragraph{Invariant Mixup.}

Recent studies~\cite{pinto2022using} show that Mixup based method leads to learning models exhibiting high entropy throughout, and consequently, Mixup method can improve the model performance on out-of-distribution data. In other words, Mixup is a good regularizer for out-of-distribution generalization.


Existing methods of Mixup on graph~\cite{han2022g,wang2021mixup,park2022graph,guo2021ifmixup} do Mixup operation on the whole graph, while we only perform Mixup method on casual subgraphs. Mixup only on invariant subgraphs enhances the performance. Applying mixup across the entire graph could potentially disrupt the real relationships. However, implementing mixup on invariant subgraphs allows for a more precise preservation and learning of original invariant relationships, reducing the occurrence of erroneous learning.  In other words, the mixed-up invariant subgraphs retain as much invariant information as possible and effectively prevent the impact of noise and spurious correlations from the entire graph on classification tasks.

We adopt Manifold Mixup on invariant subgraphs we extract in the previous part. We obtain the invariant subgraph representations $\psi^{I}_{i}$ and $\psi^{I}_{j}$ for $G_i$ and $G_j$ as:
\begin{equation}
\begin{split}
\Psi^I = \text{{GNN}}_{fea}(G^I), \psi^I = \text{{Pooling}}(\Psi^I),
\end{split}
\label{cmix1}
\end{equation}
\noindent where $\Psi^I$ is the node representation of $G^I$.

 The labels for $G_i$ and $G_j$ are $Y_j$ and $Y_j$ respectively. Our definition of invariant Mixup is as follows:
\begin{equation}
\begin{split}
\psi^{I}_{i,j} = \lambda \psi^{I}_{i} + (1 - \lambda) \psi^{I}_{j},\\
Y_{i,j} = \lambda Y_{i} + (1 - \lambda) Y_{j},\\
\lambda \sim Beta(\alpha,\alpha), \quad \quad
\end{split}
\end{equation}

\noindent where $\psi^{I}_{i,j}$ is the mixed representation of $G_{i}^I$ and $G_{j}^I$. $Y_{i,j}$ is the mixed label of $Y_i$ and $Y_j$. $\lambda$ is derived from the Beta distribution with the parameter $\alpha$. The loss function of invariant Mixup can be defined as:
\begin{equation}
\begin{split}
\mathcal{L}_{I} = \text{CrossEntropy}(Y_{ij},\hat{Y_I}),\\
\hat{Y_I} = \text{{SoftMax}}(\text{{MLP}}_{cls}(\psi_{i,j}^{I})),
\end{split}
\end{equation}

\noindent where $\hat{Y_I}$ is the predicted label with the mixed representation.

Overall, we can jointly optimize these components via the environment loss and invariant loss, i.e.,
$\mathcal{L} = \mathcal{L}_{E} + \delta \mathcal{L}_{I}$, where $\delta$ is the balance hyper-parameter.

\section{Experiments}
In this section, we conduct experiments on 11 datasets to answer the following research questions:
$\cdot$ \textbf{RQ1}: Is \model effective on the graph OOD generalization problem?
$\cdot$ \textbf{RQ2}: Is it necessary to use two kinds of Mixup?
$\cdot$ \textbf{RQ3}: How about the hyper-parameter sensitivity of \model?
$\cdot$ \textbf{RQ4}: Does the learned invariant subgraph capture invariant information, and does it capture better invariant patterns compared to other methods?



\subsection{Experimental Setup}
\paragraph{Datasets.} We conduct experiments on synthetic and real-world datasets. For synthetic datasets, following DIR~\cite{wu2022discovering}, we use the SPMotif dataset to evaluate our method on structure and degree shift. 
For real-world datasets, we examine degree shift, size shift, and other distribution shifts. For the degree shift, we employ the Graph-SST5 and Graph-Twitter datasets~\cite{chen2022learning,yuan2022explainability,dong2014adaptive,socher2013recursive}. To evaluate size shift, we utilize PROTEINS and DD datasets from TU benchmarks~\cite{morris2020tudataset}, adhering to the data split as suggested by previous research~\cite{chen2022learning}. We also consider the DrugOOD~\cite{ji2022drugood} and Open Graph Benchmark (OGB)~\cite{ogb} for structural distribution shifts. More details are shown in the Appendix.


%
\paragraph{Evaluation.} 
We employ different evaluation metrics tailored to specific datasets as previous works~\cite{chen2022learning,yang2022learning}. For the SPMotif, Graph-SST5~\cite{socher2013recursive}, and Graph-Twitter~\cite{dong2014adaptive} datasets, we use accuracy as the evaluation metric. For the DrugOOD~\cite{ji2022drugood} and OGB~\cite{ogb} datasets, we assess performance using the ROC-AUC metric. For the TU datasets~\cite{morris2020tudataset}, we measure the model with Matthews correlation coefficient. We report the mean results and standard deviations across five runs. The implementation details are given in the Appendix.
\begin{table}[b]
  \centering
\fontsize{9pt}{12pt}\selectfont
\setlength{\tabcolsep}{4pt}  
  \begin{tabular}{c|ccc} 
    \toprule[2pt]
    Dataset        & SPMotif-0.33 & SPMotif-0.6 & SPMotif-0.9 \\ 
    \midrule
    ERM            & 59.49 ± 3.50 & 55.48 ± 4.84 & 49.64 ± 4.63 \\ 
    G-mixup        & 60.31 ± 2.89 & 58.74 ± 5.58 & 53.60 ± 5.01 \\
    Manifold-mixup & 58.33 ± 4.05 & 56.63 ± 2.96 & 49.81 ± 4.25 \\ 
    \midrule
    IRM            & 57.15 ± 3.98 & 61.74 ± 1.32 & 45.68 ± 4.88 \\
    V-REx          & 54.64 ± 3.05 & 53.60 ± 3.74 & 48.86 ± 9.69 \\
    EIIL           & 56.48 ± 2.56 & 60.07 ± 4.47 & 55.79 ± 6.54 \\ 
    \midrule
    DIR            & 58.73 ± 11.9 & 48.72 ± 14.8 & 41.90 ± 9.39 \\
    GSAT           & 56.21 ± 7.08 & 55.32 ± 6.35 & 52.11 ± 7.56 \\
    CIGA           & \underline{77.33 ± 9.13} & \underline{69.29 ± 3.06} & \underline{63.41 ± 7.38} \\ 
    \midrule
    \model             & \textbf{82.36 ± 7.39} & \textbf{78.09 ± 5.63} & \textbf{76.11 ± 8.86} \\
    \bottomrule[2pt]
  \end{tabular}
  \caption{Graph classification results on synthetic datasets. We use the accuracy ACC  (\%) as the evaluation metric.}
  \label{synthetic dataset}
\end{table}

\begin{table*}[ht]
  \centering
\fontsize{9pt}{12pt}\selectfont
\setlength{\tabcolsep}{3.5pt}  
  \begin{tabular}{c|cc|cc|cccc} 
    \toprule[2pt]
    Shift Type       & \multicolumn{2}{c|}{Degree} & \multicolumn{2}{c|}{Size} & \multicolumn{4}{c}{Structure(Assay, Scaffold)} \\
    \midrule
    Dataset              & Graph-SST5 & Graph-Twitter & PROTEINS  & DD        & $\text{DrugOOD}_{\text{Assay}}$ & $\text{DrugOOD}_{\text{Scaffold}}$ & BACE & BBBP \\ 
    \midrule
     Metric       & \multicolumn{2}{c|}{ACC (\%)} & \multicolumn{2}{c|}{MCC} & \multicolumn{4}{c}{AUC (\%)} \\ 
    \midrule
     
    ERM            & 43.89 ± 1.73 & 60.81 ± 2.05 & 0.22 ± 0.09 & 0.27 ± 0.09 & 76.41 ± 0.73 & 66.83 ± 0.93 & 77.83 ± 3.49 & 66.93 ± 2.31 \\ 
    G-Mixup        & 43.75 ± 1.34 & 63.91 ± 3.01 & 0.24 ± 0.03 & 0.29 ± 0.04 & 76.53 ± 2.20 & 66.01 ± 1.35 & 79.12 ± 2.75 & 68.44 ± 2.08 \\
    Manifold-Mixup & 43.11 ± 0.65 & 62.60 ± 1.87 & 0.23 ± 0.04 & 0.28 ± 0.06 & \underline{77.02 ± 1.15} & 65.56 ± 0.44 & 78.85 ± 1.26 & 68.67 ± 1.38 \\ 
    \midrule
    IRM            & 43.69 ± 1.26 & 63.50 ± 1.23 & 0.21 ± 0.09 & 0.22 ± 0.08 & 74.03 ± 0.58 & 66.32 ± 0.27 & 77.51 ± 2.46 & 69.13 ± 1.45 \\
    V-REx          & 43.28 ± 0.52 & 63.21 ± 1.57 & 0.22 ± 0.06 & 0.21 ± 0.07 & 75.85 ± 0.78 & 65.37 ± 0.42 & 76.96 ± 1.88 & 64.86 ± 2.13 \\
    EIIL           & 42.98 ± 1.03 & 62.76 ± 1.72 & 0.20 ± 0.05 & 0.23 ± 0.10 & 76.93 ± 1.44 & 64.13 ± 0.89 & 79.36 ± 2.72 & 65.77 ± 3.36 \\ 
    \midrule
    DIR            & 41.12 ± 1.96 & 59.85 ± 2.98 & 0.25 ± 0.14 & 0.20 ± 0.10 & 74.11 ± 3.10 & 64.45 ± 1.69 & 79.93 ± 2.03 & \underline{69.73 ± 1.54} \\
    GSAT           & 43.72 ± 0.87 & 62.50 ± 1.44 & 0.21 ± 0.06 & 0.28 ± 0.04 & 76.64 ± 2.82 & 66.02 ± 1.13 & 79.63 ± 1.87 & 68.48 ± 2.01 \\
    CIGA           & \underline{44.71 ± 1.14} & \underline{64.45 ± 1.99} & \underline{0.40 ± 0.06} & \underline{0.29 ± 0.08} & 76.15 ± 1.21 & \underline{67.11 ± 0.33} & \underline{80.98 ± 1.25} & 69.65 ± 1.32 \\ 
    \midrule
    \model         & \textbf{46.69 ± 0.52} & \textbf{66.23 ± 1.58} & \textbf{0.43 ± 0.05} & \textbf{0.36 ± 0.04} & \textbf{78.16 ± 0.65} & \textbf{68.32 ± 0.48} & \textbf{82.65 ± 1.17} & \textbf{71.03 ± 0.79} \\
    \bottomrule[2pt]
  \end{tabular}
  \caption{Graph OOD generalization performance with on real-world datasets. 
We show the graph classification results on datasets with three types of distribution shifts: degree, size, and structure. We use ACC (\%) as the evaluation metric on Graph-SST5 and Graph-Twitter datasets, MCC for DD and PROTEIN datasets, and ROC-AUC (\%) for $\text{DrugOOD}_{\text{scaffold}}$, $\text{DrugOOD}_{\text{assay}}$, BACE, and BBBP datasets. Experimental results indicate that our method outperforms all the baselines.}
  \label{real-world datasets}
\end{table*}

\paragraph{Baselines.} 
In addition to Empirical Risk Minimization (ERM), we compare our approach with three categories of methods, including mixup-based, invariant learning based and graph OOD generalization methods. In the mixup-based methods, there are Manifold Mixup~\cite{verma2019manifold} and G-Mixup~\cite{han2022g}. For invariant learning based category,  we consider the methods that use known environment partition such as Invariant Risk Minimization (IRM)~\cite{arjovsky2019invariant} and V-REx~\cite{krueger2021out}, as well as methods that automatically partition environments, such as EIIL~\cite{creager2021environment}. The third category encompasses those information bottleneck based methods like CIGA~\cite{chen2022learning} and GSAT~\cite{miao2022interpretable}, as well as methods based on environment divisions within the graph, such as DIR~\cite{wu2022discovering}.

\subsection{Main Results (RQ1)}
\paragraph{Experiments on synthetic datasets.}
We report our results on synthetic datasets in Table \ref{synthetic dataset}. 
The bias which set as 0.33, 0.6, and 0.9 represents the degree of spurious correlation between labels and features.
Through the experimental results, we can observe that our proposed method outperforms other baselines by a large margin under three different bias settings. Specifically, our model surpasses ERM by an average margin of 44.2\% on average and outperforms the state-of-the-art (SOTA) method CIGA by 13.1\% on average. This demonstrates that our IGM is more adept at capturing the invariant patterns under distribution shifts, thereby enabling the model to perform better on OOD data.

\begin{figure*}[h]
    \centering
    \begin{subfigure}[t]{0.23\textwidth}
        \raisebox{-\height}{\includegraphics[width=\textwidth]{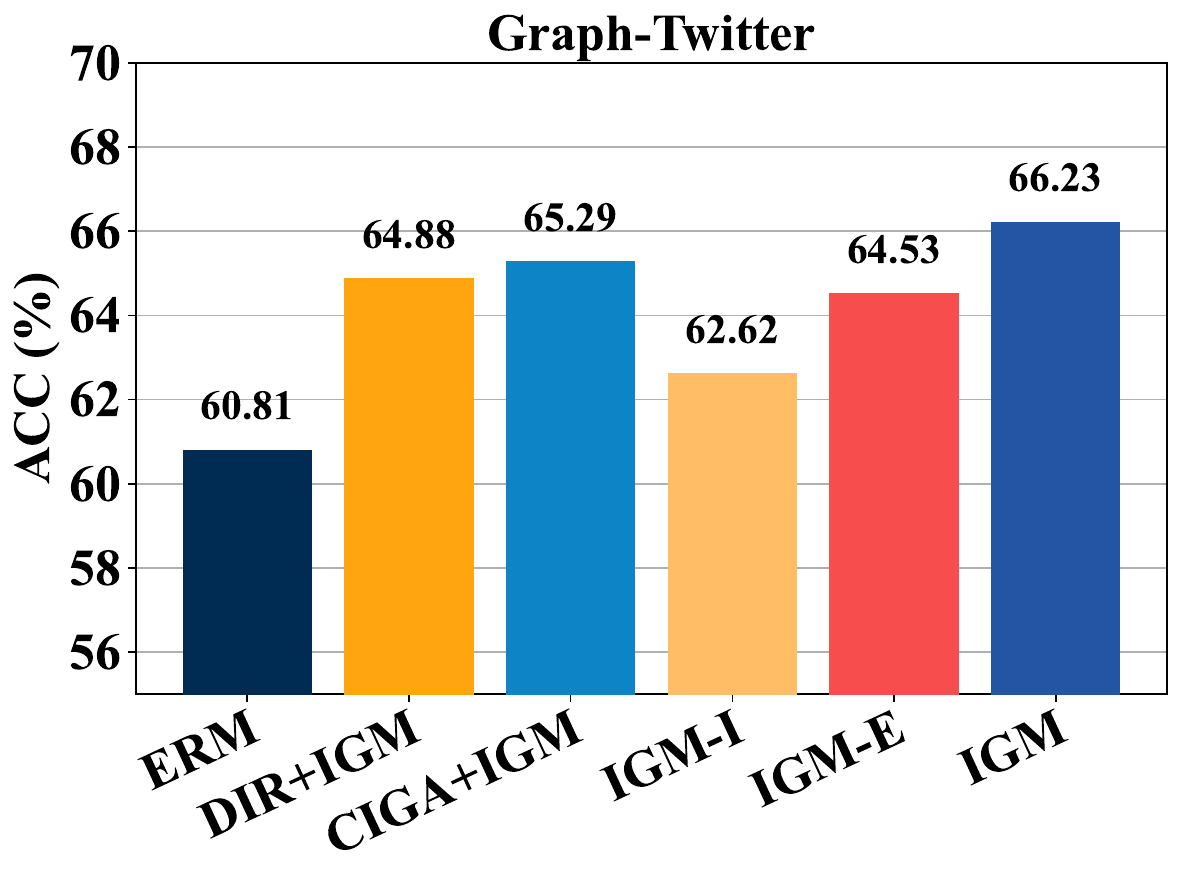}}
    \end{subfigure}
    \hfill
    \begin{subfigure}[t]{0.23\textwidth}
        \raisebox{-\height}{\includegraphics[width=\textwidth]{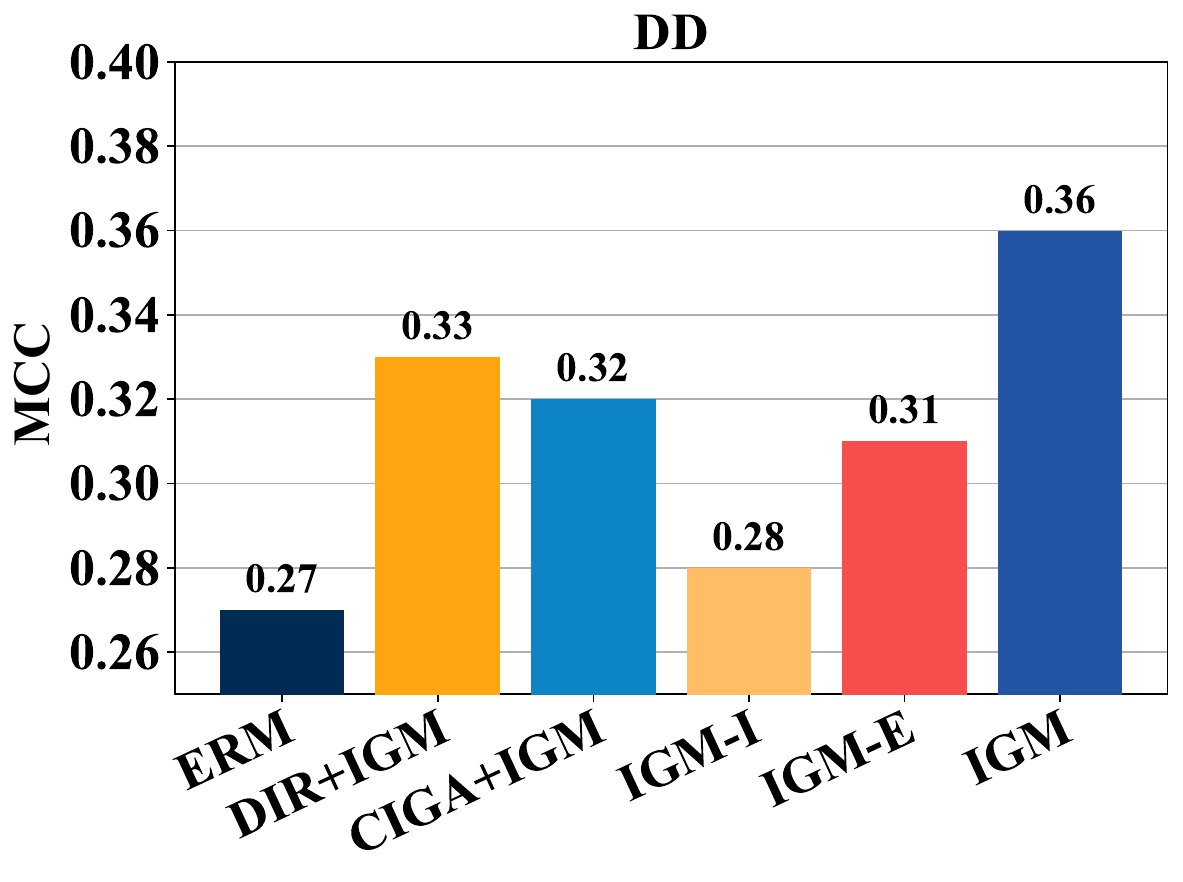}}

    \end{subfigure}
    \hfill
    \begin{subfigure}[t]{0.23\textwidth}
        \raisebox{-\height}{\includegraphics[width=\textwidth]{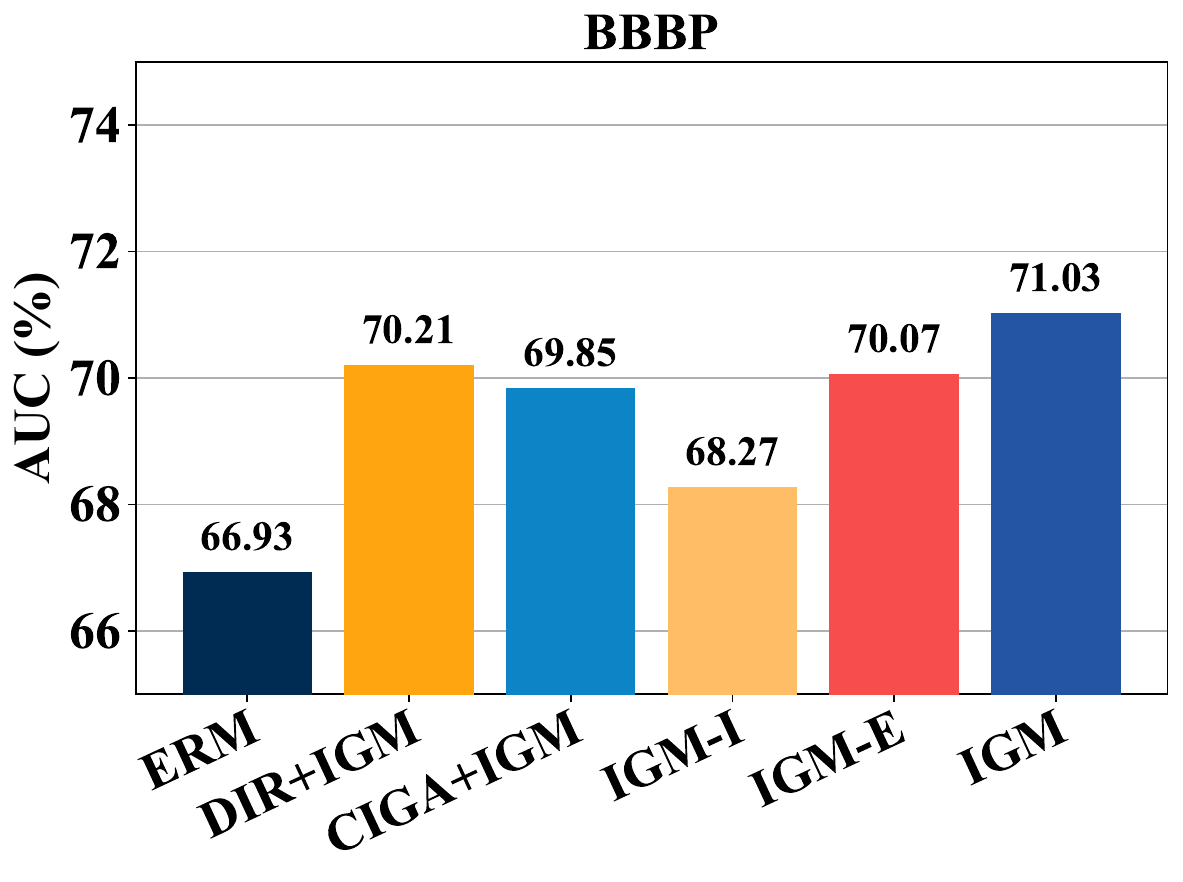}}

    \end{subfigure}
    \hfill
    \begin{subfigure}[t]{0.23\textwidth}
        \raisebox{-\height}{\includegraphics[width=\textwidth]{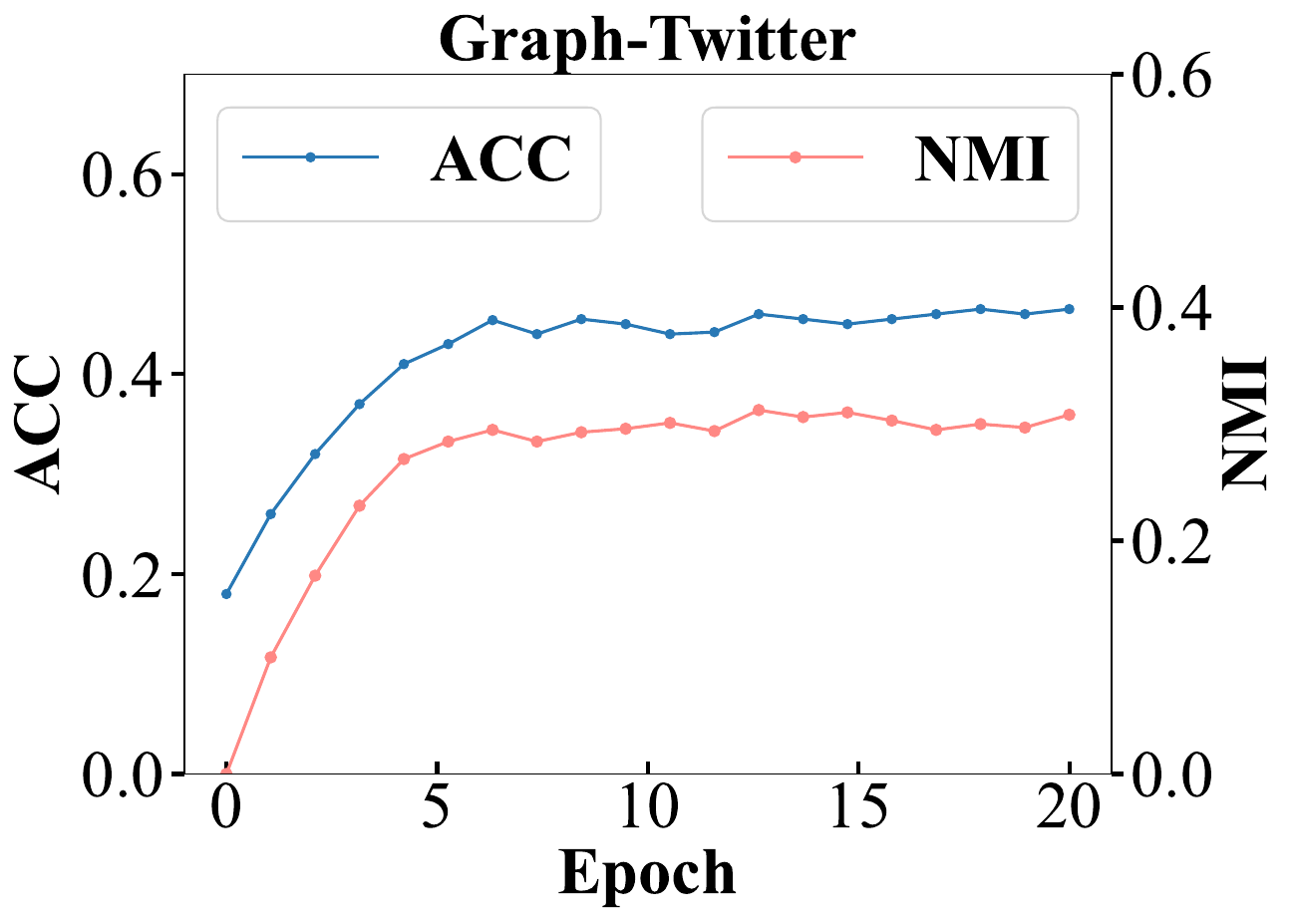}}

    \end{subfigure}
    \caption{The first three figures present ablation studies on Graph-Twitter, DD, and BBBP datasets, emphasizing the importance of utilizing both environment and invariant Mixup. The last figure provides an analysis of environment subgraph clustering and the result of classification, demonstrating the mutual enhancement of the two mixup components.}
        \label{ablation}
\end{figure*}

\paragraph{Experiments on real-world datasets.}
%

We explore three types of distribution shifts on real-world datasets: degree shift, size shift, and structure shift. The results are presented in Table \ref{real-world datasets}. It can be observed that existing methods uniformly fail to achieve good OOD generalization across all datasets. For instance, G-Mixup underperforms compared to ERM on the Graph-SST5 and NCI109 datasets, while IRM is consistently outdone by ERM on most datasets, and the SOTA method CIGA is outperformed by ERM on DrugOOD-Scaffold dataset.

As can be observed across these eight datasets, our model consistently achieved the best performance. Our method demonstrates an overall improvement of 7.7\% compared to the state-of-the-art (SOTA) methods. These results show that our model can effectively deal with the complex distribution shift in the real world. It also indicates the model's strong OOD generalization ability.

In detail, for datasets with size shift, our method achieves an average enhancement of 15.9\% over SOTA. In instances involving degree shift, the average improvement stands at 3.9\%. For datasets subjected to structure shift, our method records an average increase in performance of 1.7\%. From these results, it can be inferred that our model excels in identifying invariant patterns in all these distribution shifts.

\subsection{Ablation Study (RQ2)}
We conduct two types of experiments for the ablation study. First, we explore the necessity of using two mixup methods to find invariant subgraphs. Second, we investigate the contributions of each component of the proposed \model. For the first part, we initially utilize a previous OOD graph generalization method (CIGA, DIR) for training. We then use the subgraph extractor from the trained model as our model's invariant subgraph extractor and fix its parameters. Then we train with our two kinds of mixup. For the second part, we compare our model (two kinds of mixup) with two ablated models (only invariant Mixup and only environment Mixup) and ERM. We conduct experiments on the Graph-Twitter, DD, and BBBP datasets, obtaining results under three different distribution shifts. The results are demonstrated in Figure \ref{ablation}.

We can observe that the performance of using the subgraph extractor from the previous methods (CIGA, DIR) combined with our two mixup methods for training is superior to ERM but still falls short of our method. Specifically, the IGM outperformed the IGM using a pre-trained extractor by an average of 4.6\%. This validates the necessity of employing both mixup methods to obtain invariant subgraphs. Furthermore, we can observe that the performance of the two ablated models are somewhat diminished compared to simultaneously using both mixup methods, yet they still outperform ERM by 1.6\% on average. This indicates that both types of mixup can achieve OOD generalization to a certain extent. Among them, the model trained only with invariant Mixup shows a more significant reduction in performance than the model using only environment Mixup. 

\paragraph{Collaboration of Two Mixups.}
To show the environment Mixup module and invariant Mixup module can be mutually promoted by each other, we record the test accuracy and the Normalized Mutual Information (NMI)~\cite{strehl2002cluster}, which is a common clustering metric and can reflect the quality of generated environments for invariant learning. 
As shown in Figure~\ref{ablation}, the results on Graph-Twitter demonstrate that such two  metrics improve synchronously during the training process. One plausible reason is that, during environment Mixup, invariant learning across different environments captures invariant information, thereby promoting the invariant Mixup. Conversely, in the invariant Mixup phase, the invariant information captured amplifies the environment Mixup by delineating environments with larger distribution shifts, subsequently enhancing invariant learning in the environment Mixup segment.


\subsection{Hyper-parameter Sensitivity Analysis (RQ3)}

\begin{figure}[h]
    \centering
    \includegraphics[width=0.22\textwidth]{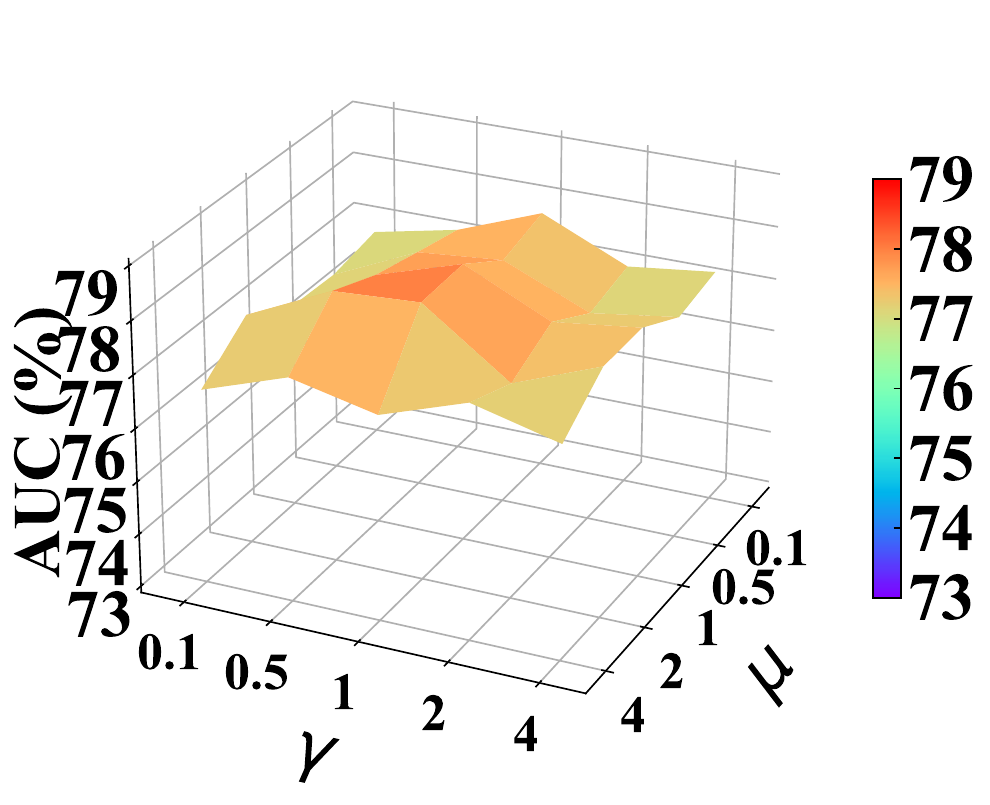} 
    \hfill
    \includegraphics[width=0.22\textwidth]{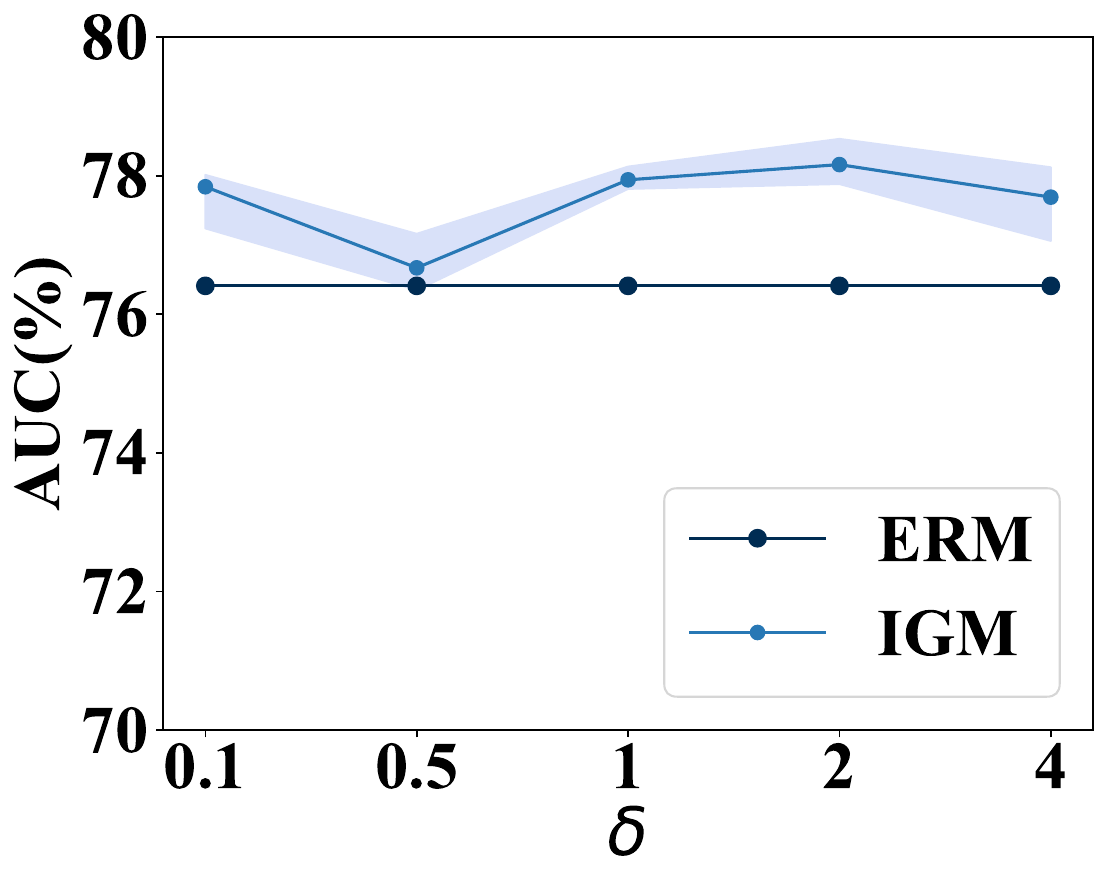} 
    \caption{AUC sensitivity of hyper-parameters $\gamma$, $\mu$, $\delta$}
    \label{hpyer}
\end{figure}

We conduct experiments on DrugOOD to examine our model's sensitivity to hyper-parameters. We select three critical parameters of the model, including the IRM weight $\gamma$, V-REx weight $\mu$, invariant Mixup weight $\delta$. We vary $\gamma$, $\mu$ and $\delta$ in $\{0.1, 0.5, 1, 2, 4\}$. The results are shown in Figure~\ref{hpyer}.
Our method remains stable and effective across different values of these hyper-parameters.
\\

\begin{figure}[h]
    \centering
    
    \begin{subfigure}[b]{0.15\textwidth}
        \includegraphics[width=\textwidth]{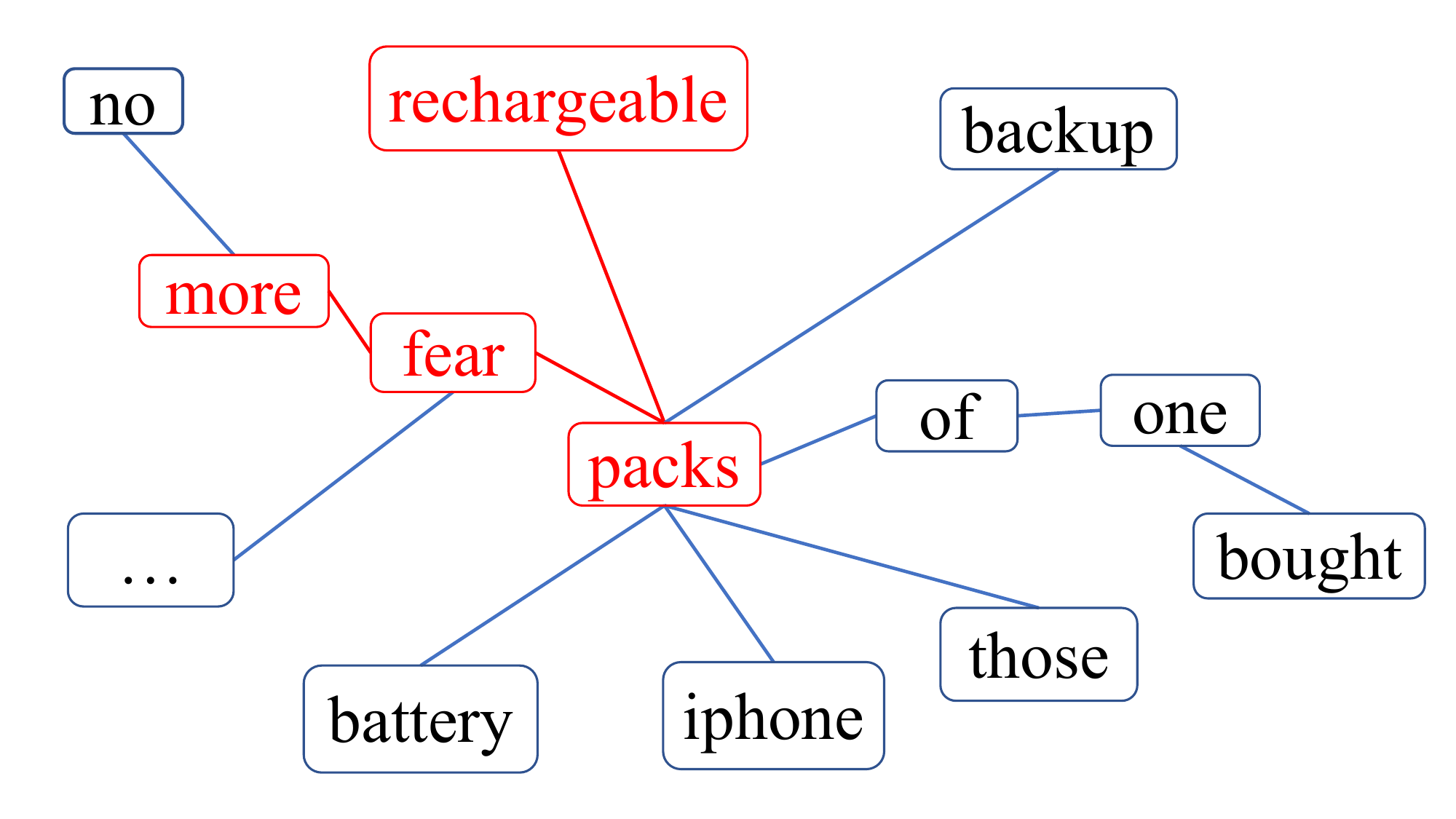}
        \caption{DIR}
    \end{subfigure}
    \hfill
    \begin{subfigure}[b]{0.15\textwidth}
        \includegraphics[width=\textwidth]{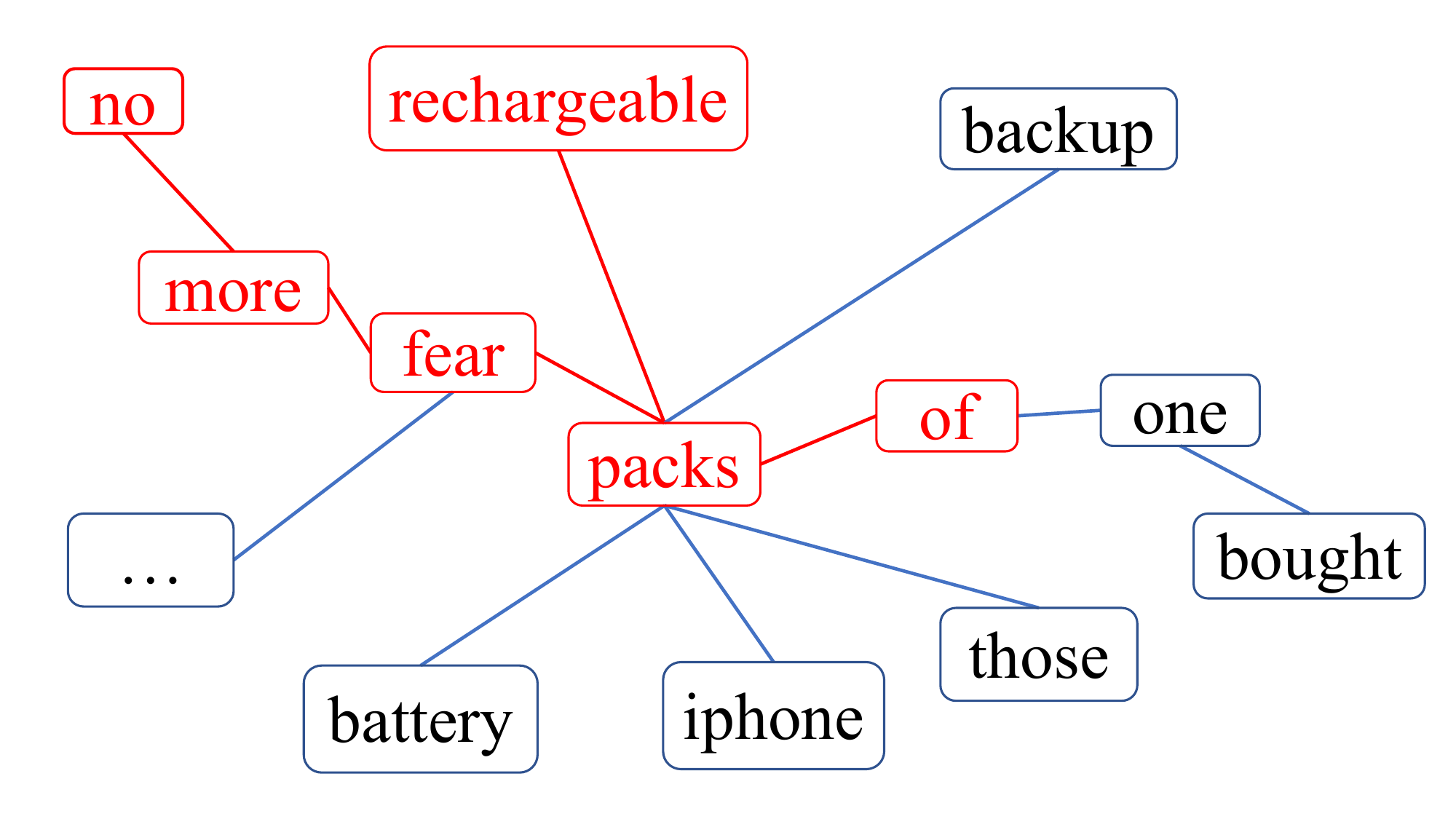}
        \caption{CIGA}
    \end{subfigure}
    \hfill
    \begin{subfigure}[b]{0.15\textwidth}
        \includegraphics[width=\textwidth]{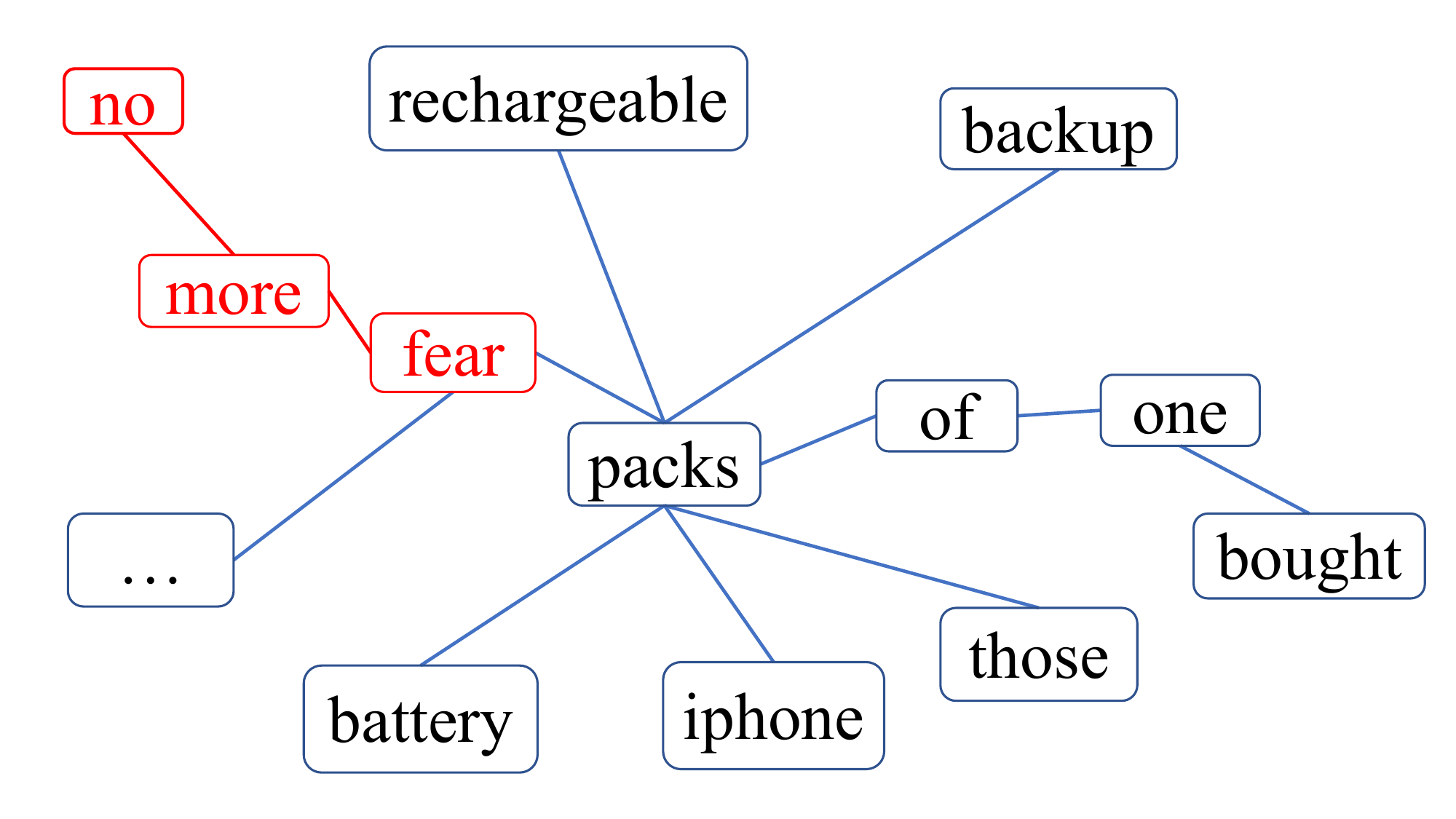}
        \caption{IGM}
    \end{subfigure}
    
    \caption{Visualization of the invariant subgraphs extracted by different models on Graph-SST dataset. The original sentence is ``Bought one of those rechargeable iPhone backup battery packs... no more fear".}
\end{figure}



    

\subsection{Invariant Subgraph Visualization (RQ4)}
To verify whether our method captures the invariant information, we first visualize the invariant subgraphs found by our model and other graph OOD methods on the Graph-Twitter, and we use this dataset because it is comprehensible to humans. It can be observed that our model adeptly identifies the specific subgraphs that are pivotal in determining the sentiment of the sentences. In contrast, DIR fails to capture all the proper subgraphs, resulting in classification errors, while CIGA tends to capture relatively larger subgraphs. 

\section{Conclusion}

In this work, we integrate mixup with invariant learning to address the problem of OOD generalization in graphs. We propose two modules, namely environment Mixup and invariant Mixup, to capture invariant information within graphs, thereby achieving OOD generalization. Extensive experiments demonstrate the efficacy of our methods under various distribution shifts on both synthetic and real-world datasets. In future work, we aim to extend our framework to node classification tasks and explore its applicability to dynamic graphs.

\label{conclusion}

\section{Acknowledgments}
This work is supported in part by the National Natural Science Foundation of China (No. U20B2045, 62192784, U22B2038, 62002029, 62322203, 62172052), Young Elite Scientists Sponsorship Program (No. 2023QNRC001) by CAST.

\bibliography{aaai24}

\end{document}